\documentclass[jou,apacite]{apa6}

\usepackage[utf8]{inputenc}
\usepackage[T1]{fontenc}

\usepackage{amsmath}
\usepackage{amsfonts}
\usepackage{graphicx}
\usepackage{caption}
\usepackage{subcaption}

\usepackage{microtype}
\usepackage{algorithm}
\usepackage{algpseudocode}
\usepackage{siunitx}
\usepackage{booktabs}

\usepackage{multirow}

\title{Unsupervised domain adaptation for \\ medical imaging segmentation with self-ensembling}
\shorttitle{Unsupervised domain adaptation for medical imaging segmentation with self-ensembling}

\fourauthors{Christian S. Perone}{Pedro Ballester}{Rodrigo C. Barros}{Julien Cohen-Adad}
\fouraffiliations{NeuroPoly Lab, Institute of Biomedical Engineering, \\ Polytechnique Montreal, Montreal, QC, Canada.}{Machine Intelligence and Robotics Research Group \\ School of Technology \\ Pontifícia Universidade Católica do Rio Grande do Sul \\ Porto Alegre, RS, Brazil}{Machine Intelligence and Robotics Research Group \\ School of Technology \\ Pontifícia Universidade Católica do Rio Grande do Sul \\ Porto Alegre, RS, Brazil}{NeuroPoly Lab, Institute of Biomedical Engineering, \\ Polytechnique Montreal, Montreal, QC, Canada. \\ Functional Neuroimaging Unit, CRIUGM, \\ Universite de Montreal, Montreal, QC, Canada.}

\abstract{Recent advances in deep learning methods have come to define the state-of-the-art for many medical imaging applications, surpassing even human judgment in several tasks. Those models, however, when trained to reduce the empirical risk on a single domain, fail to generalize when applied to other domains, a very common scenario in medical imaging due to the variability of images and anatomical structures, even across the same imaging modality. In this work, we extend the method of unsupervised domain adaptation using self-ensembling for the semantic segmentation task and explore multiple facets of the method on a small and realistic publicly-available magnetic resonance (MRI)  dataset. Through an extensive evaluation, we show that self-ensembling can indeed improve the generalization of the models even when using a small amount of unlabelled data.}

\rightheader{Unsupervised domain adaptation for medical imaging segmentation with self-ensembling}
\leftheader{Christian S. Perone, Pedro Ballester, Rodrigo C. Barros, Julien Cohen-Adad}

\begin{document}
\maketitle    
                        
\section{Introduction}
In the past few years, the research community has witnessed the fast developmental pace of deep learning~\cite{LeCun2015a} approaches for unstructured data analysis, arguably establishing an important scientific milestone. Deep neural networks constitute a paradigm shift from traditional machine learning approaches for unstructured data. Whereas the latter rely on hand-crafted feature engineering for improving learning over images, text, audio, and similarly unstructured inputs, deep neural networks are capable of automatically learning robust hierarchical features, in what is known as \textit{representation learning}. Deep learning approaches have achieved human-level performance on many tasks and, indeed, sometimes even surpassing it in applications such as natural image classification~\cite{He2015b}, or arrhythmia detection~\cite{Rajpurkar2017}.

Due to its popularity and compelling results in many domains, deep learning attracted a lot of attention from the medical imaging community. A recent survey by Litjens~et~al.~\cite{Litjens2017} analyzed more than 300 medical imaging studies, and found that deep neural networks have become pervasive throughout the field of medical imaging, with a significant increase in the number of publications between 2015 and 2016. The survey also identified that the most addressed task is image segmentation, likely due to the importance of quantification of anatomical structures and pathologies~\cite{Gros2018} for disease diagnosis and prognosis, as opposed to less informative tasks such as classification of pathologies or detection of structures, which can be posed as a segmentation tasks as well, but not the opposite.

Deep neural networks are thus becoming the norm in medical imaging, though there are still several unsolved challenges that remain to be addressed. For instance, one of the most well-known problems is the high sample complexity, or how much data deep learning requires to accurately learn and perform well on unseen images, which is related to the concepts of model complexity and generalization, active areas of research in learning theory~\cite{neyshabur2017exploring}.

The large amount of required data to train deep neural networks can be partially mitigated with techniques such as transfer learning~\cite{Yosinski2014,Zamir_2018_CVPR}. However, transfer learning is problematic in medical imaging because a large dataset is still required so the models can benefit from the inductive transfer process. Unlike the case of natural images, where annotations can be easily provided by non-experts, medical images require careful and time-consuming analysis from trained experts such as radiologists.

Yet another challenge when deploying deep learning models to medical imaging analysis -- and perhaps one of the most difficult to solve -- is the so-called \textit{data distribution shift}, wherein different imaging scenarios (e.g. parameter choices, different protocols) can result in vastly different data distributions, despite imaging a common object. Therefore, models trained under the empirical risk minimization (ERM) principle, might fail to generalize to other domains due to its strong assumptions. ERM is the statistical learning principle behind many machine learning methods, and it offers good learning guarantees and bounds if its assumptions hold, such as the fact that the training and test datasets derive from similar domains. However, in practice, this assumption is often violated.

\begin{figure}[!htpb]
	\centering
	\includegraphics[width=0.97\linewidth]{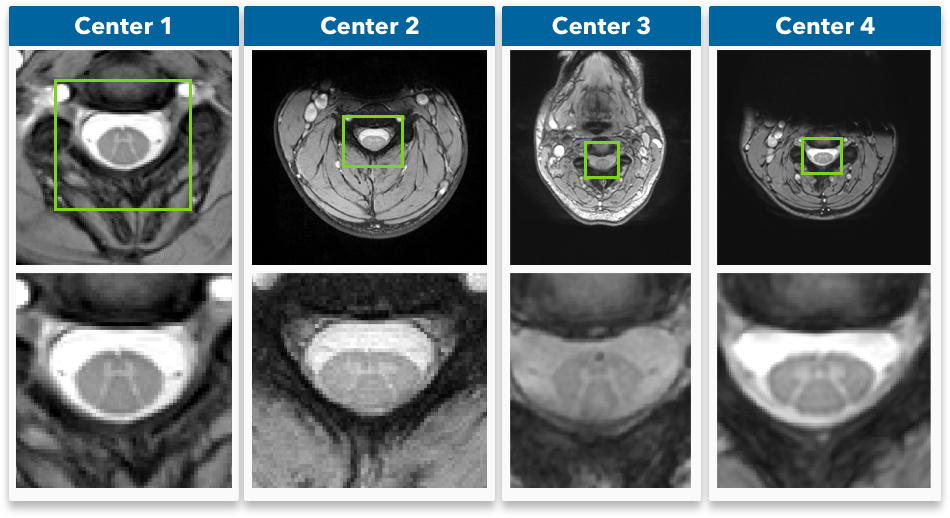}
	\caption{\label{fig:variability-sample}
		Samples of axial MRI from four different centers (UCL, Montreal, Zurich, Vanderbilt) that participated in the SCGM Segmentation Challenge~\protect\cite{Prados2017}. \textbf{Top row}: original MRI images. \textbf{Bottom row}: crop of the spinal cord (green rectangle). Reproduced from~\protect\cite{Perone2018}. Best viewed in color.
	}
\end{figure}

\begin{figure}[!htpb]
	\centering
	\includegraphics[width=0.97\linewidth]{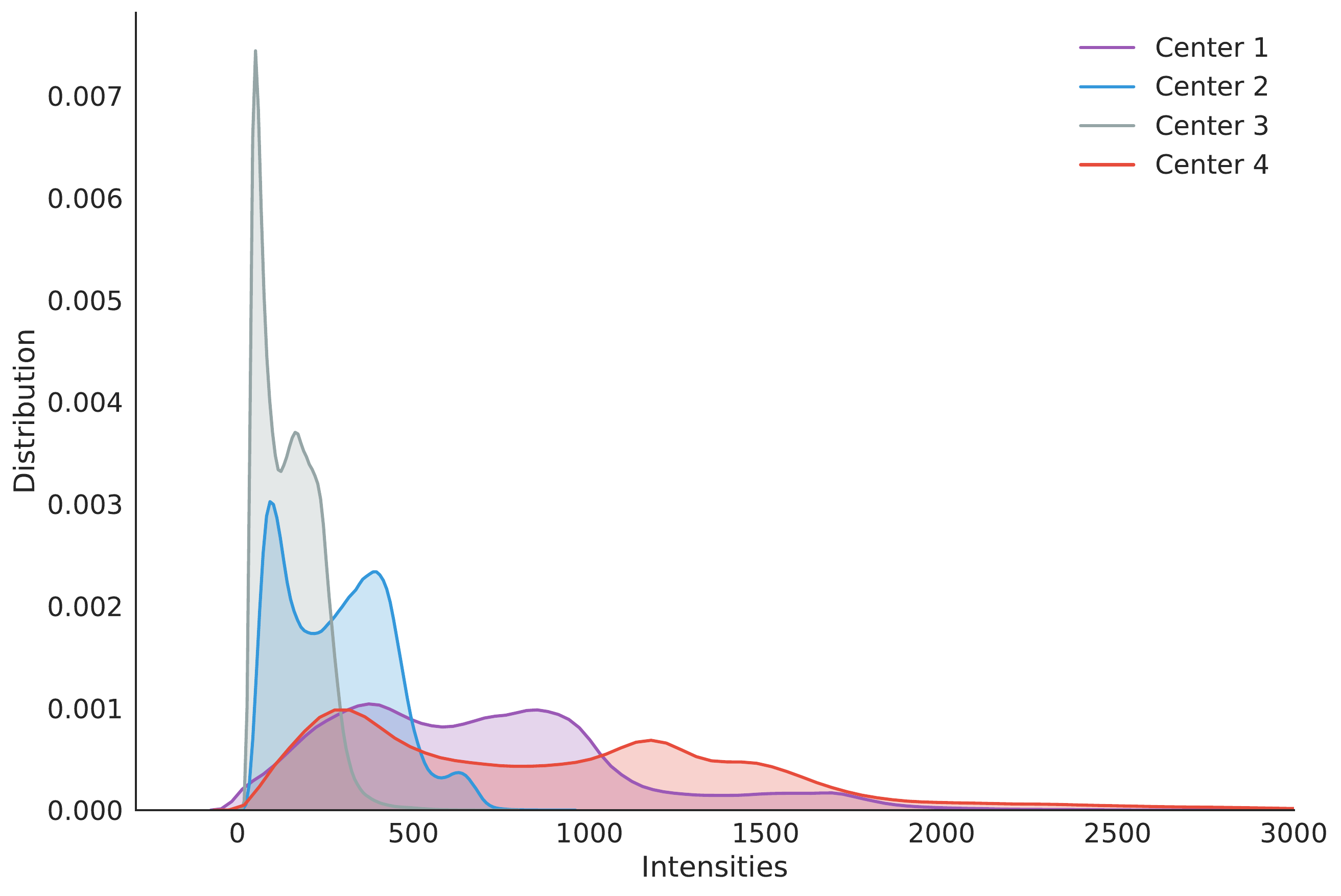}
	\caption{\label{fig:distribution}
		MRI axial-slice pixel intensity distribution from four different centers (UCL, Montreal, Zurich, Vanderbilt) that collaborated to the SCGM Segmentation Challenge~\protect\cite{Prados2017}.
	}
\end{figure}

When a deep learning model that assumes independent and identically-distributed (iid) data is trained with images from one domain and is subsequently deployed on images from a different domain (e.g. distinct imaging center), that follow a distinct data distribution, its performance often degrades by a large margin. 
An example of domain shift can be seen in magnetic resonance imaging (MRI), where the same machine vendor, using the same protocol, and for the same subject, can nevertheless produce different images. Variability tends to be even greater between different centers where machine vendor, software versions, radio-frequency coils and sequence parameters (e.g., slice positioning, image resolution) often vary. Figure~\ref{fig:variability-sample} illustrates those inter-center differences in data distribution, based on data from the Gray Matter (GM) segmentation challenge~\cite{Prados2017}. Figure~\ref{fig:distribution} illustrates the associated voxel intensity distribution for the same dataset.

Although this distribution shift is common in medical imaging, the problem is surprisingly ignored during the design of many different challenges in the field. It is common to have the same domain data (same machine, protocol, etc.) on both training and test sets. However, this homogeneous data split often does not represent the reality and in many cases will produce over-optimistic evaluation results.
In practice, it is rare to have labeled data available from a new center before training a model, hence it is common to use a pre-trained model from a different domain on completely different data. Therefore, it is paramount to have a proper evaluation avoid contaminating the test set with data from the same domain that is present in the training set. Incurring the risk of the detrimental effects of inadequate evaluations~\cite{Zech2018}. The name given to learn a classifier model or any other predictor with a shift between the training and the target/test distributions is known as ``domain adaptation'' (DA). In this work we expand upon a previously-developed method~\cite{french2017self} for DA based on the Mean Teacher~\cite{tarvainen2017mean} approach, to segmentation tasks, the most addressed task in medical imaging.

We provide the following contributions: (i) we extend the unsupervised DA method using self-ensembling for the semantic segmentation task; to the best of our knowledge, this is the first time this method is used for semantic segmentation in medical imaging; (ii) we explore some model components such as different consistency losses, and evaluate the performance of our method on a series of experiments using a realistic small MRI dataset; (iii) we perform an ablation experiment to provide strong evidence that unlabeled data is responsible for the observed performance improvement, ruling out the effects of the exponential moving average; (iv) we provide visualizations to derive insight into the model dynamics of the unsupervised DA task.

This paper is organized as follows. In Section~\ref{sec:related-work} we present related work, in Section~\ref{sec:semi-supervised} we give a brief treatment to the unsupervised DA task and its connection to semi-supervised learning. In Section~\ref{sec:method} we describe our method in terms of model architecture and corresponding design decisions. In Section~\ref{sec:materials} we describe the dataset used in our experiments and how we performed the data split for the DA scenario. In Section~\ref{sec:experiments} we provide the experiment results, followed by an ablation study in Section~\ref{sec:ablation-experiments}. In Section~\ref{sec:visualization} we provide visual insights from the adaptation dynamics of the model for multiple domains. Finally, in Section~\ref{sec:conclusion-limitation} we discuss our findings and limitations of our work. In the spirit of open science and reproducibility, we also provide more information regarding data and source-code availability in Section~\ref{sec:source-code}.

\section{Related work}
\label{sec:related-work}
Deep learning methods for medical imaging has become a popular research focus in recent years~\cite{Litjens2017}. Before the development of deep learning models, initial work was focused mostly on patch-based~\cite{coupe2011patch} segmentation.
With the growing interest in deep learning for computer vision, the first attempts using Convolutional Neural Networks (CNNs) for image segmentation processed image patches through a sliding window, to yield segmented patches, which were then stitched together to yield the final segmented image~\cite{lai2015deep}. The main drawbacks of this approach are computational cost (i.e., several forward passes are required to produce the segmented images) and inconsistency in predictions, the latter of which can be fixed or partially mitigated by overlapping sliding windows, depending on the network architecture.

Though patch-wise methods continue to be actively researched~\cite{hou2016patch} and have led to several advances in segmentation~\cite{lai2015deep}, presently, the most common deep architecture for segmentation is or is based on the so-called Fully Convolutional Network (FCN)~\cite{long2015fully}. This architecture is solely based on convolutional layers with the final result not depending on the use of fully-connected layers. FCNs can provide a fully-segmented image within a single forward step, and with variable output size depending on the size of the input tensor. One of the most well-known FCNs for medical imaging is the U-net~\cite{Ronneberger2015}, which combines convolutional, downsampling, and upsampling operations with skip non-residual connections. In this work we used the U-Net architecture, although the proposed framework is decoupled from the choice of network architecture, as further discussed in Section~\ref{sec:network}.

Deep Domain Adaptation (DDA), which is a field unrelated in essence to medical imaging, has been widely studied in the recent years~\cite{wang2018deep}. We can divide the literature on DDA as follows: (i)~methods based on building domain-invariant feature spaces through auto-encoders~\cite{ghifary2016deep}, adversarial training~\cite{ganin2016domain}, GANs~\cite{hoffman2017cycada,Sankaranarayanan2018}, or disentanglement strategies~\cite{Liu2018, cao2018dida}; (ii) methods based on the analysis of higher-order statistics~\cite{li2016revisiting,sun2016deep}; (iii) methods based on explicit discrepancy between source and target domains~\cite{tzeng2014deep}; and (iv)~self-ensembling methods based on implicit discrepancy~\cite{french2017self,tarvainen2017mean}. 

In~\cite{hoffman2017cycada}, the authors trained GANs with cycle-consistent loss functions~\cite{zhu2017unpaired} to remap the distribution from the source to the target dataset, thereby creating target domain specific features for completing the task. In~\cite{Sankaranarayanan2018}, GANs were employed as a means of learning aligned embeddings for both domains. Similarly, disentangled representations for each domain have been proposed~\cite{Liu2018,cao2018dida} with the goal of generating a feature space capable of separating domain-dependent and domain-invariant information. 

In \cite{li2016revisiting}, the authors proposed to change parameters of the neural network layers for adapting domains by directly computing or optimizing higher-order statistics. More specifically, they proposed an alternative for batch normalization called Adaptive Batch Normalization (AdaBN) that computes different statistics for the source and target domains, hence creating domain-invariant features that are normalized according to the respective domain. In a similar fashion, Deep CORAL~\cite{sun2016deep} provides a loss function for minimizing the covariances between target and source domain features.

Discrepancy-based methods pose a different approach to DDA. By directly minimizing the discrepancy between activations from the source and target domain, the network learns to generate reasonable predictions while incorporating information from the target domain. The seminal work of Tzeng et al.~\cite{tzeng2014deep} directly minimizes the discrepancy between a specific layer with labeled samples from the source set and unlabeled samples from the target set. 

Implicit discrepancy-based methods such as self-ensembling~\cite{french2017self} have become widely used for unsupervised domain adaptation. Self-ensembling is based on the Mean Teacher network~\cite{tarvainen2017mean}, which was first introduced for semi-supervised learning tasks. Due to the similarity between unsupervised domain adaptation and semi-supervised learning, there are very few adjustments that need to be made to employ the method for the purposes of DDA. Mean Teacher optimizes a task loss and a consistency loss, the latter minimizing the discrepancy between predictions on the source and target dataset. We further detail how Mean Teacher works in Section~\ref{sec:mean-teacher}. 

There are a few studies that report results of using different data domains for medical imaging by making use of the unsupervised domain adaptation literature. The work~\cite{albadawy2018deep} discusses the impact of deep learning models across different institutions, showing a statistically significant performance decrease in cross-institutional train-and-test protocols. A few studies have applied domain adaptation to medical imaging directly by using adversarial training~\cite{kamnitsas2017unsupervised,Chen2018,Zhang2018,Lafarge2017,Javanmardi2018,Dou2018}, with some studies using generative models to augment training~\cite{Mahmood2018,Madani2018}. Nevertheless, to the best of our knowledge, this present work is the first to address the problem of domain shift in medical image segmentation by extending the unsupervised DA self-ensembling method to semantic segmentation tasks.

\section{Semi-supervised learning and \\ unsupervised domain adaptation}
\label{sec:semi-supervised}

A common approach for improving training when few labeled examples are available is semi-supervised learning, which is defined as follows: given a labeled dataset with distribution $P(X_l)$ and unlabeled data with distribution $P(X_u)$, learn from both labeled and unlabeled data in order to improve a supervised learning task (say, classification) or an unsupervised learning task (say, clustering).

Semi-supervised learning methods tend to perform well when unlabeled data actually come from the same distribution as the labeled data. This allows the learning algorithm to leverage its knowledge using unlabeled data, which usually represents the majority of samples. As promising as semi-supervised learning is, the assumption that the distribution of unlabeled data $P(X_u)$ is similar to $P(X_l)$ often fails in real-world applications. We refer the reader to a thorough evaluation of semi-supervised learning methods and their limitations in~\cite{odena2018realistic}.

It often happens that models are applied in situations that are largely different from those in which they were originally trained. Examples include different weather conditions for outdoor activity recognition, or different cities for training driverless vehicles. Those changes in scenario shift the data distribution $P(X)$, reducing the quality of the predictions in cases where the model was not properly adapted to the desired condition.

The difference between the distributions from the examples used in training and test sets is called \textit{domain shift}. Consider a source dataset with input distribution $P(X_s)$ and label distribution $P(Y | X_s)$, as well as a target dataset with input distribution $P(X_t)$ and labels $P(Y | X_t)$, $P(X_s) \neq P(X_t)$. Domain adaptation can be addressed via a supervised approach where labeled data from the target domain is available, or via unsupervised learning where only unlabeled data is available for the target domain. 

When a method addresses the problem of domain adaptation using unlabeled data for the target domain, which is the most common and useful scenario, the task at hand is called \textit{unsupervised domain adaptation}. Unsupervised domain adaptation methods assume that distributions $P(X_s)$, $P(Y|X_s)$ and $P(X_t)$ are available, while $P(Y | X_t)$ is not. In other words, only the source dataset provides labeled examples. Hence, the task is to leverage knowledge from the target domain using the unlabeled data available in $P(X_t)$.

\section{Method}
\label{sec:method}
This section details the base domain adaptation methods that we used for the medical image application. We further discuss the changes that are needed to enable unsupervised domain adaptation for segmentation tasks, as opposed to the typical classification scenario.

\subsection{Self-ensembling and mean teacher}
\label{sec:mean-teacher}
Self-ensembling was originally conceived as a viable strategy for generating predictions on unlabeled data~\cite{laine2016temporal}. The original paper proposes two different models for self-ensembling. The first model, called $\Pi$, employs a consistency loss between predictions on the same input. Each input from a batch is passed twice through a neural network, each time with distinct augmentation parameters, to yield two different predictions. A squared difference between those predictions is minimized along with the cross-entropy for the labeled examples. The second model, called \emph{temporal ensembling}, works under the assumption that as the training progresses, averaging the predictions over time on unlabeled samples may contribute to a better approximation of the true labels. This pseudo-label is then considered as a target during training. The squared difference between the averaged predictions and the current one is minimized along with the cross-entropy for labeled examples. The network performs the exponential moving average (EMA) to update the generated targets at every epoch:

\begin{equation}\label{eq:ema}
    f'(x)_t = \alpha f'(x)_{t-1} + (1-\alpha) f(x)_t
\end{equation}

Where $t$ is the step, $x$ is the data, $f(\cdot)$ is the network and $\alpha$ is a momentum term that controls how far the ensemble reaches training history data.

Self-ensembling was extended to directly combine model weights instead of predictions. This adaptation is called the Mean Teacher~\cite{tarvainen2017mean} model. Considering Eq.~(\ref{eq:ema}) for updating the target pseudo-labels, Mean Teacher updates the model weights at each step, thus generating a slightly improved model compared to the model without the EMA, a framework which is linked to the Polyak-Ruppert Averaging~\cite{polyak1992acceleration, ruppert1988efficient}. In this scenario, the EMA model was named teacher, and the standard model, student. The update function is as follows:

\begin{equation}
    \theta'_t = \alpha\theta'_{t-1} + (1-\alpha) \theta_t
\end{equation}

where $\theta$ are the model parameters, $t$ is the step and $\alpha$ is the hyperparameter regulating the importance of the current model's weights with respect to previous models. The best results are achieved when $\alpha$ is increased later on during training, causing the model to forget more about the parameters during earlier stages of training than later when the network is performing better.

Each training step involves a loss component for both labeled and unlabeled data. All samples from a batch are evaluated by both the student and teacher models, with their respective predictions compared via the consistency loss. The labeled data, however, is also compared to its ground truth, as traditionally performed in segmentation tasks, in what we call the task loss:

\begin{equation}
    J(\theta) = J_{task}(\theta) + \gamma J_{consistency}(\theta) + \lambda R(\theta)
\end{equation}
where $\gamma$ and $\lambda$ are the Lagrange multipliers that represent, respectively, the consistency and regularization weights. The $\gamma$ hyperparameter was empirically found to improve results when it varied through time, given that in the earlier training steps the network continues to generate poor results. The consistency weight follows a sigmoid ramp-up saturating at a given user-defined value.

Mean Teacher follows the dynamics of model distillation~\cite{hinton2015distilling}. In this scenario, a trained model is used for predicting instances and its output is used as labels for another, smaller model. This is considered a good practice as soft labels tend to better represent the characteristics of the classes (e.g., the representation distance between a Siberian Husky and an Alaskan Malamute should arguably be smaller than the distance between a Siberian Husky and a Persian Cat). Unlike traditional distillation formulations, the Mean Teacher framework also uses the teacher model to generate labels for unlabeled data and represents a model of the same size that is simultaneously updated during training.

The Mean Teacher framework was also extended for unsupervised domain adaptation in~\cite{french2017self}. Among the proposed changes, the authors modified the data batches such that each batch consists of images from both the source and target domains. At each step, the student model evaluates images from the source domain and computes derivatives via a task loss based on the ground truth. The target domain images, which are unlabeled, are used to compute the consistency loss by comparing predictions from both student and teacher models. It differs from its original formulation in that the teacher model only has access to unlabeled examples (in this case, examples from the target domain). Each loss function is thus responsible for improving learning at a single domain. The task loss is evaluated by comparing the predictions against the ground truth for the labeled examples (source domain). For the consistency loss, MSE is often used to evaluate the predictions from both student and teacher models for the unlabeled examples (target domain).

\subsection{Adapting mean teacher for segmentation tasks}

Both the original and adapted Mean Teacher versions for unsupervised domain adaptation rely on the cross-entropy classification cost. Given that we are not dealing with classification but with a segmentation task, we need to minimize a different loss function that takes into consideration the particularities of that task. Originally proposed in~\cite{milletari2016v}, the Dice loss generates reliable segmentation predictions due to its low sensitivity to class imbalance:

\begin{equation}
    J_{task}(\theta) = - \frac{2*\sum_{i}^{N}p_ig_i}{\sum_{i}^{N}p_i + \sum_{i}^{N}g_i}
    \label{eq:dice}
\end{equation}
where $p_i$ and $g_i$ are flattened predictions and ground truth values for an instance, respectively. Dice was kept as the task loss for both baseline and adaptation experiments. Note that the dice loss is computed for the entire batch at once, unlike the typical strategy of averaging when using cross-entropy, for instance.

\begin{figure*}[!htpb]
	\centering
	\includegraphics[width=0.97\linewidth]{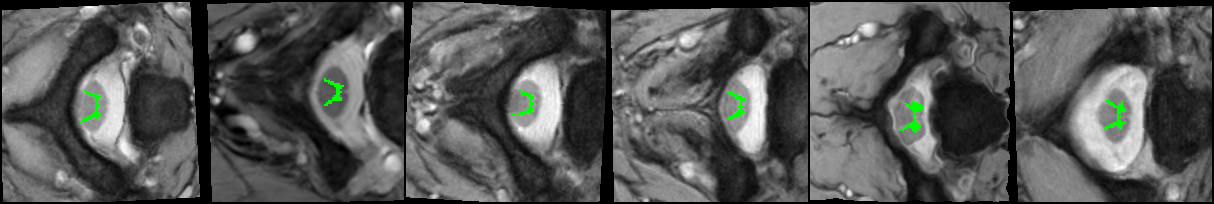}
	\caption{\label{fig:ground_truth}
		Data augmentation result of random MRI axial-slices samples from the SCGM Segmentation Challenge~\protect\cite{Prados2017}. 
		The ground truth is shown in green with the same transformation parameters applied. Best viewed in color.
	}
\end{figure*}

\begin{figure}[!htpb]
	\centering
	\includegraphics[width=1.0\linewidth]{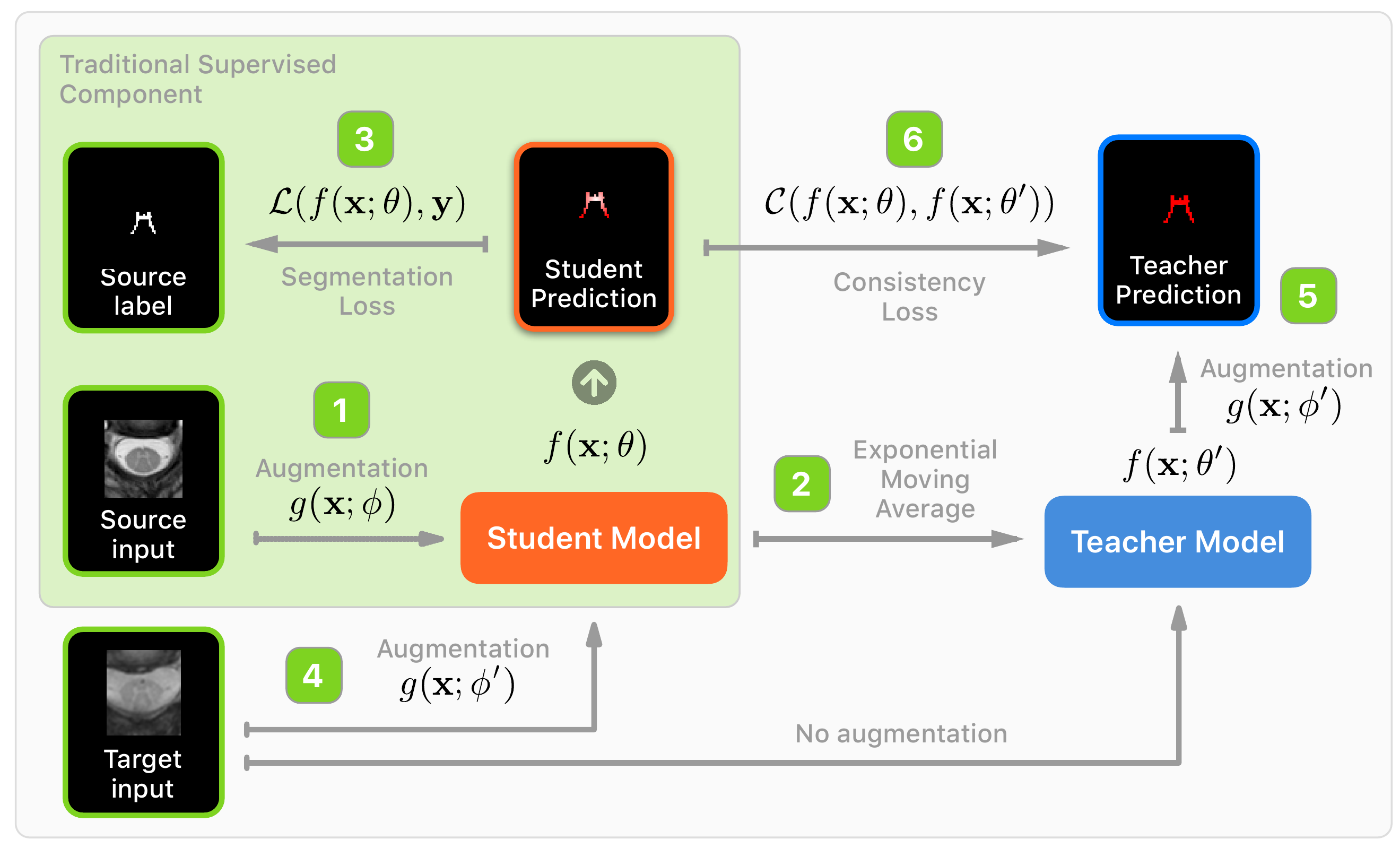}
	\caption{\label{fig:algorithm_flow} Overview of the proposed method. The green panel represents the traditional supervision framework.
	\textbf{(1)}~The source domain input data is augmented by the $g(\bm{x}; \phi)$ transformation and fed into the student model.
	\textbf{(2)}~The teacher model parameters is updated with an exponential moving average (EMA) from the student weights.
	\textbf{(3)}~The traditional segmentation loss, where the supervision signal is provided with the source domain labels.
	\textbf{(4)}~The input unlabeled data from the target domain is transformed with $g(\bm{x};\phi^\prime)$ before the student model forward pass (note the different parametrization $\phi^\prime$).
	\textbf{(5)}~The teacher model prediction is transformed with $g(\bm{x};\phi^\prime)$ (same transformation as in Step 4).
	\textbf{(6)}~The consistency loss, which enforces consistency between student and teacher predictions.
	}
\end{figure}

\begin{figure}[!htpb]
	\centering
	\includegraphics[width=0.97\linewidth]{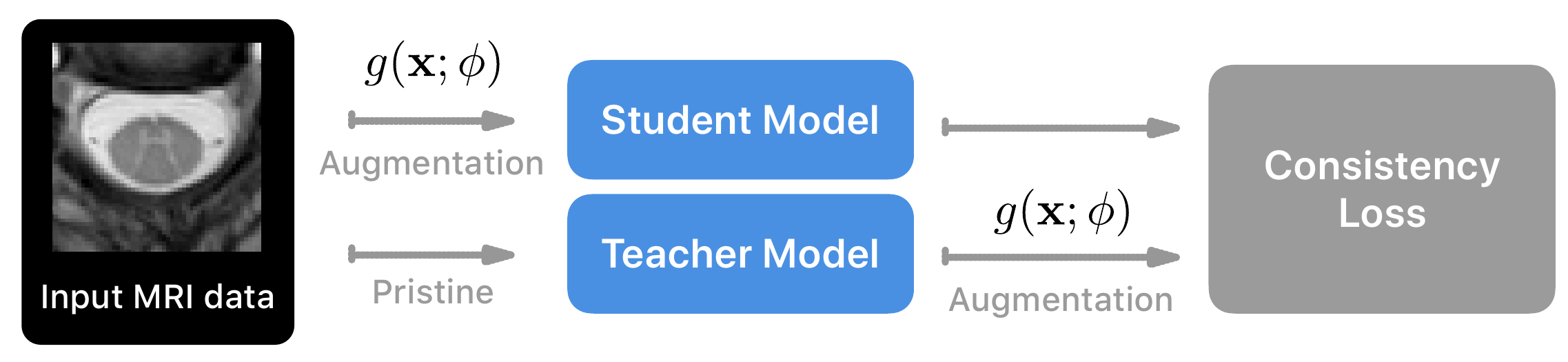}
	\caption{\label{fig:data-augmentation}
		Data augmentation scheme used to overcome the spatial misalignment between student and teacher model predictions. The same augmentation parameters are used for the input data for the student model and on the teacher model predictions.
	}
\end{figure}

A second problem when training the student and teacher models for segmentation tasks is the inconsistency introduced between training samples of the student and teacher models when a spatial transformation (e.g., translation, rotation, or any similar spatial transformation for the purpose of data augmentation) is applied with different parameters to both inputs of the teacher and student models. To solve that problem we used the same approach employed by~\cite{Perone2018a} as shown in Figure~\ref{fig:data-augmentation}. The spatial transformation $g(x; \phi)$, where $x$ is the input data and $\phi$ are the transformation parameters (i.e., rotation angle), is applied to the student model before feeding data into the model. For the teacher model, the same transformation $g(x; \phi)$ is applied to the predictions of the teacher model, causing both predictions to be aligned for the consistency loss. This framework is possible because backpropagation only occurs for the student model and therefore there is no need for differentiation on the delayed augmentation of the teacher model. The proposed method is illustrated in Figure~\ref{fig:algorithm_flow}. Examples of images after data augmentation and their respective compensated ground truth are shown in Figure~\ref{fig:ground_truth}. 

\subsection{Model architecture}
\label{sec:network}

Since the U-net~\cite{Ronneberger2015} is widely applied in medical imaging field for diverse tasks, in order to provide results that can generalize to a wide spectrum of applications, for all experiments we employed the U-net~\cite{Ronneberger2015} model architecture with 15 layers, group normalization~\cite{Wu2018}, and dropout. The rationale behind group normalization and not batch normalization is discussed later.

To provide a fair comparison, we followed the recommendations from~\cite{Oliver2018} and kept the same model for the baseline and for our method. While the Mean Teacher model also acts as a regularizer, we kept the same regularization weights for all comparisons. Regularization weights can be fine-tuned, however, possibly improving even further the results of Mean Teacher.

\subsection{Baseline employed}
\label{sec:baseline}
We conducted an extensive hyperparameter search to find a proper baseline model, yielding a mini-batch size of 12 and a dropout rate of 0.5. For training, we used the Adam optimizer~\cite{Kingma2015a} with $L_2$ penalty factor of $\lambda = \num{6e-4}$, $\beta_1 = 0.99$, and $\beta_2 = 0.999$. For learning rate, we used a sigmoid learning rate ramp-up strategy until epoch 50 followed by a cosine ramp-down until epoch 350. Eq.~(\ref{eq:sigmoid_rampup}) shows the sigmoid ramp-up strategy:

\begin{equation}
    \label{eq:sigmoid_rampup}
    R_{up}(m) = \alpha e^{-5(1-m)^2}
\end{equation}
where $\alpha$ is the highest learning rate and $m$ represents the ratio between current epoch and the total ramp-up epochs. Eq.~(\ref{eq:cosine_rampdown}) presents the cosine ramp-down strategy:

\begin{equation}
    \label{eq:cosine_rampdown}
    R_{down}(r) = \alpha \frac{\cos(\pi r) + 1}{2}
\end{equation}
where $\alpha$ is the highest learning rate and $r$ is the ratio between the number of epochs after the ramp-up procedure and the total number of epochs expected for training. 

For a fair comparison, and to be able to assess the specific benefits of domain adaptation, no hyperparameter from the baseline model was changed in the adaptation scenario. The only change concerned the hyperparameters, which only affect the domain adaptation training procedure. 

\subsection{Consistency loss}
The consistency loss is one of the most important aspects of Mean Teacher. If the difference between predictions from teacher and student is not representative enough for distilling the knowledge on the student model, the method will not properly work or training may even diverge. In the original implementation of the Mean Teacher method, the mean squared error (MSE) was proposed:

\begin{equation}
    J_{MSE}(\theta) = \frac{\sum_{i}^{N}(p_i - g_i)^2}{N}
    \label{eq:mse}
\end{equation}
where $p_i$ and $g_i$ are flattened predictions from student and teacher, respectively.

As an alternative, the cross-entropy is more commonly used for classification tasks. The cross-entropy is defined as:

\begin{equation}
    J_{CE}(\theta) = - \sum_{i}^{N}p_i\log{g_i}
    \label{eq:cross_entropy}
\end{equation}

where $p_i$ and $g_i$ are predictions from student and teacher, respectively. However, cross-entropy is also known to be sensitive to class imbalance.

Our preliminary experiments led to use MSE with different weights per class to address the problem of class imbalance. However, this approach relies on thresholding predictions from the teacher to define binary expected voxel values for the student. Defining both the correct weights and the threshold value is a difficult task that did not seem to improve overall results.

The same problem happens with more complex losses, e.g., the Focal Loss~\cite{lin2018focal}, due to additional hyperparameters (in this case, $\gamma$ and $\beta$).

We have thus explored other losses: the Dice loss, presented in Section~\ref{sec:method}, and the Tversky loss~\cite{salehi2017tversky}. The Tversky loss is a variation of the dice loss that aims at mitigating the problem of class imbalance, which is common in medical image segmentation tasks. It is defined as:

\begin{equation}
    J_{tversky}(\theta) = - \frac{\sum_{i}^{N}p_{0i}g_{0i}}{\sum_{i}^{N}p_{0i}g_{0i} + \alpha\sum_{i}^{N}p_{0i}g_{1i} + \beta\sum_{i}^{N}p_{1i}g_{0i}}
    \label{eq:tversky}
\end{equation}
where $p_{0i}$ and $g_{0i}$ represent the predicted probabilities and expected ground-truth of a voxel that belongs to the correct tissue, whereas $p_{1i}$ and $g_{1i}$ respectively represent the predicted probabilities and expected ground-truth (0 or 1) of a voxel that belongs to any other tissue. The $\alpha$ and $\beta$ hyperparameters address the problem of class imbalance. The Tversky loss, however, is hampered by the difficulty of determining more hyperparameters alongside the consistency weight value (same issue as noted above with the weighted MSE).

We have also noticed that both Dice and Tversky coefficients are problematic when used as consistency losses. Albeit properly representing the nature of the task, their formulation is based on multiplication and it is assumed that the ground-truth is binary, i.e. $g_i \in \{0, 1\}$. However, given that we use the teacher soft outputs (i.e., not binary), both Dice and Tversky losses do not obey the proper score orientation: $S(G, y) > S(G^*, y)$, where $S$ is the scoring function and $y$ is the ground truth. This relationship should hold only if $G$ is a better probabilistic forecast, which is not the case for Tversky and Dice when using soft targets.

For example, if $p_i = 0.9$ and $g_i = 1.0$, the numerator yields $0.9$. However, when $p_i = 0.9$ and $g_i = 0.9$, the score should increase (because the predicted and ground-truth are the same), but instead the numerator decreases to $0.81$ and the output score also decreases.

One way to overcome this issue is to threshold the teacher's predictions such that the loss functions can work as expected. However, identifying suitable threshold values is not trivial since they drastically impact how the network adapts, and reduces the benefits of using a distillation-based~\cite{hinton2015distilling} approach. An alternative to thresholding is to modify the formulations of the loss functions such that they can properly handle non-binary labels. A detailed analysis of such modifications falls outside the scope of this paper so we left it for future work.

\subsection{Batch Normalization and Group Normalization \\ for domain adaptation}
Batch Normalization~\cite{Ioffe2015} is a method used to improve the training of deep neural networks through the stabilization of the distribution of layer inputs. Nowadays, Batch Normalization is pervasive in most deep learning architectures, enabling the use of large learning rates and helping with convergence.

Initially thought to help with the \emph{internal covariate shift} (ICS) problem~\cite{Ioffe2015}, Batch Normalization was recently found~\cite{Santurkar2018} to smooth the optimization landscape of the network due to the improvement of the Lipschitzness, or $\beta$-smoothness ~\cite{Santurkar2018} of both loss and gradients.

Batch Normalization works differently for training and inference. During training, the normalization happens using the batch statistics, while on inference it uses the population statistics, usually estimated with moving averages on each batch during the training procedure. This strategy, however, is problematic for domain adaptation via Mean Teacher, given that there are multiple distributions being fed during training, causing the Batch Normalization statistics to be computed with both source and target data.

One possible approach to overcome that issue is to use different batch statistics for the source and the target domains as done in AdaBN~\cite{li2016revisiting}. Implementing this approach within the training procedure is easily achieved using modern frameworks because it only requires to forward the batch to each domain separately~\cite{french2017self}. However, in the implementation of French et al., both source and target domain data were used to compute the running average at inference. One should ideally perform running averages and population statistics on both domains separately, though at the expense of increased complexity on training, especially when running on a multi-GPU scenario with small batch sizes, a very common scenario in segmentation tasks where synchronization is also required.

Besides the mentioned issues, Batch Normalization also suffers from sub-optimal results when using small batch sizes~\cite{Wu2018}, which are very common in segmentation tasks due to memory requirements. For those reasons, we chose Group Normalization~\cite{Wu2018}, an alternative to Batch Normalization where channels are divided into groups and where mean and variance are computed within each group regardless of batch sizes. Group Normalization works consistently better than Batch Normalization with small batch sizes (typically <15) and does not require storing running averages for the population statistics, simplifying the training and inference procedures and providing better results for our scenario that involves domain adaptation and segmentation tasks.

\subsection{Hyperparameters for unsupervised \\ domain adaptation}
A problem shared by many techniques for unsupervised domain adaptation is how to set proper hyperparameters such as the learning rate or the consistency weight. In unsupervised settings, there are no labeled data from the target domain so the estimation of hyperparameters from the source distribution alone can be completely different from those from the target distribution.

An alternative method to solve this issue is to use \emph{reverse cross-validation}~\cite{Zhong}, which was also used in~\cite{ganin2016domain}. However, once again, this approach comes at the expense of increasing the complexity of the validation process. Nevertheless, we found that the estimation of hyperparameters for Mean Teacher on the source domain yielded robust results, therefore we adopted them in our experiments. We are aware that such a simple strategy is a limitation of our evaluation procedure since we could probably achieve better results for our proposed method by incorporating a more sophisticated hyperparameter estimation procedure.

\section{Materials}
\label{sec:materials}

The Spinal Cord Gray Matter Challenge \cite{Prados2017} dataset is a multi-center, multi-vendor, and publicly-available MRI data collection that is comprised of 80 healthy subjects with 20 subjects from each center.

The demographics of the dataset range from a mean age of 28.3 up to 44.3 years old. Three different MRI systems were employed (Philips Achieva, Siemens Trio, Siemens Skyra) with distinct acquisition parameters. The voxel size resolution of the dataset ranges from $0.25\times0.25\times2.5$ mm up to $0.5\times0.5\times5.0$ mm and the number of axial slices ranged from 3 to 28. The dataset is split between training (40) and test (40) sets, and the test set labels are hidden (not publicly available). For each labeled slice in the dataset, 4 gold-standard segmentation masks were manually created by 4 independent experts (one per participating center). For more detailed information regarding the dataset (e.g., the MRI parameters), please see~\cite{Prados2017}.

Since the Spinal Cord Gray Matter Challenge dataset contains data from all 4 centers both in the training and test sets, we used a non-standard split in order to evaluate our technique within the domain adaptation scenario, where the domain present in the test set is not contaminated by the training data domain. Therefore, we used centers 1 and 2 as the training set, center 3 as the validation set, and center 4 as the test set. 

\begin{figure}
	\centering
	\includegraphics[width=0.7\linewidth]{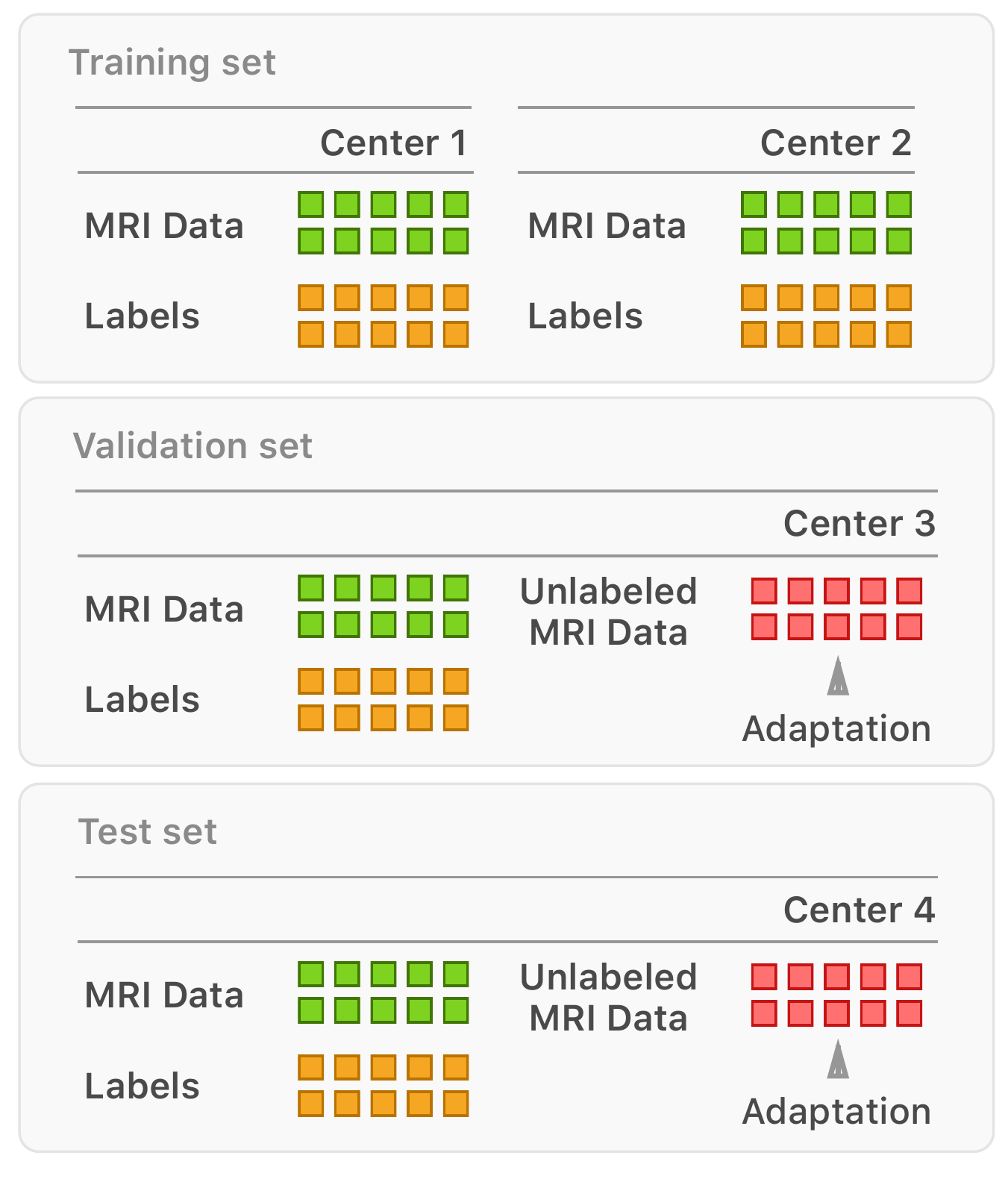}
	\caption{\label{fig:data-split}
		Overview of the data splitting method for training machine learning models. Each colored square represents a single subject of the dataset (containing multiple axial slices).
	}
\end{figure}

We used the unlabeled data from center 4 test set (which does not contain publicly-available labels) as the unlabeled data for the target domain, and we used the training data from center 4 (with labels) as the test set to evaluate the final performance of our model. We also slice all 3D samples into 2D axial slices and resampled each slice to $0.25\times0.25$ mm. An overview of the dataset is presented in Figure~\ref{fig:data-split}. 

\section{Experiments}
\label{sec:experiments}
We have designed several experiments to understand the behavior of different aspects of domain adaptation on the medical imaging domain. We have also performed ablation studies and evaluated multiple metrics for each center.

\subsection{Adapting to different centers}

\begin{table*}[!htpb]
    \centering
    \caption{Evaluation results in different centers. The evaluation and adaptation columns represent, respectively, the centers where testing and adaptation data were collected. Results are averages and standard deviations over 10 runs (with independent initialization of random weights). Values highlighted represent the best results at each center. All experiments were trained in both centers 1 and 2 simultaneously. Dice represents the Sørensen–Dice coefficient and mIoU represents the mean Intersection over Union.}
    \label{tab:results}
    \begin{tabular}{cccccccc}
        \toprule
        \textbf{Evaluation} & \textbf{Adaptation} & \textbf{Dice} & \textbf{mIoU} & \textbf{Recall} & \textbf{Precision} & \textbf{Specificity} & \textbf{Hausdorff} \\
         \midrule

         \multirow{3}{*}{Center 1} & Baseline & 47.25 $\pm$ 0.10 & 31.46 $\pm$ 0.08 & \textbf{94.90} $\pm$ 0.29 & 32.18 $\pm$ 0.09 & 99.66 $\pm$ 0.0 & 2.88 $\pm$ 0.01 \\
         & Center 3 & 47.71 $\pm$ 0.16 & 31.84 $\pm$ 0.14 & 94.18 $\pm$ 0.16 & 32.69 $\pm$ 0.15 & 99.67 $\pm$ 0.0 & \textbf{2.85} $\pm$ 0.01 \\
         & Center 4 & \textbf{48.42} $\pm$ 0.92 & \textbf{32.47} $\pm$ 0.80 & 94.51 $\pm$ 0.57 & \textbf{33.33} $\pm$ 0.93 & \textbf{99.68} $\pm$ 0.02 & 2.86 $\pm$ 0.02 \\ 
        
         \midrule
        
         \multirow{3}{*}{Center 2} & Baseline & 50.69 $\pm$ 0.09 & 34.44 $\pm$ 0.08 & \textbf{94.79} $\pm$ 0.24 & 35.32 $\pm$ 0.10 & 99.61 $\pm$ 0.00 & 2.89 $\pm$ 0.01 \\
         & Center 3 & 51.05 $\pm$ 0.25 & 34.76 $\pm$ 0.23 & 93.78 $\pm$ 0.42 & 35.83 $\pm$ 0.31 & 99.62 $\pm$ 0.01 & \textbf{2.87} $\pm$ 0.01 \\
         & Center 4 & \textbf{51.29} $\pm$ 0.67 & \textbf{34.98} $\pm$ 0.61 & 93.87 $\pm$ 0.91 & \textbf{36.06} $\pm$ 0.82 & \textbf{99.63} $\pm$ 0.02 & 2.87 $\pm$ 0.02 \\ 
        
        \midrule

        \multirow{3}{*}{Center 3} & Baseline & 82.81 $\pm$ 0.33 & 71.05 $\pm$ 0.36 & \textbf{90.61} $\pm$ 0.63 & 77.09 $\pm$ 0.34 & 99.86 $\pm$ 0.0 & 2.14 $\pm$ 0.02 \\
        & Center 3 & \textbf{84.72} $\pm$ 0.18 & \textbf{73.67} $\pm$ 0.28 & 87.43 $\pm$ 1.90 & \textbf{83.17} $\pm$ 1.62 & \textbf{99.91} $\pm$ 0.01 & \textbf{2.01} $\pm$ 0.03 \\
        & Center 4 & 84.45 $\pm$ 0.14 & 73.30 $\pm$ 0.19 & 87.13 $\pm$ 1.77 & 82.92 $\pm$ 1.76 & \textbf{99.91} $\pm$ 0.01 & 2.02 $\pm$ 0.03 \\ 
        
        \midrule
        
        \multirow{3}{*}{Center 4} & Baseline & 69.41 $\pm$ 0.27 & 53.89 $\pm$ 0.31 & \textbf{97.22} $\pm$ 0.11 & 54.95 $\pm$ 0.35 & 99.70 $\pm$ 0.00 & 2.50 $\pm$ 0.01 \\
        & Center 3 & 73.27 $\pm$ 1.29 & 58.50 $\pm$ 1.57 & 94.92 $\pm$ 1.48 & 60.93 $\pm$ 2.51 & 99.77 $\pm$ 0.03 & 2.36 $\pm$ 0.06 \\
        & Center 4 & \textbf{74.67} $\pm$ 1.03 & \textbf{60.22} $\pm$ 1.24 & 93.33 $\pm$ 1.96 & \textbf{63.62} $\pm$ 2.42 & \textbf{99.80} $\pm$ 0.02 & \textbf{2.29} $\pm$ 0.05
        \\
        \bottomrule

    \end{tabular}
\end{table*}

We trained the network with both centers 1 and 2 in a supervised fashion. We then adapted the network to centers 3 and 4 separately. With this setup, we were able to address three related research questions on adaptation and semi-supervised learning:
\begin{enumerate}
    \item How do predictions change at inference time when images from domains different than the source domain are presented?
    \item How does the network change its predictions to the novel domain after performing domain adaptation?
    \item How well does an adapted network generalize when presented with images that were not used during training, neither as a supervised signal nor as an unsupervised adaptation component?
\end{enumerate}

Results of this first experiment are presented in Table~\ref{tab:results}. 

Regarding \textit{Question 1}. Both centers 1 and 2 are included in the training set and we would like to assess whether additional unsupervised data from different domains (centers 3 or 4) improve generalization on the centers 1 and 2. For both adapted centers 3 and 4, results for all metrics (except for recall) outperform the baseline, suggesting a positive change in prediction performance for the source domain after domain adaptation on unseen domains leveraging unlabelled data.

To answer \textit{Question 2}, one can analyze the rows where both evaluation and adaptation centers are the same (3 or 4). Both rows present the highest values for almost all metrics (again, excepted for recall). This suggests that domain adaptation is working properly for that scenario.

Regarding \textit{Question 3}, by looking at evaluation on center 3 and adaptation using center 4 (and vice-versa), we observe gains over the baseline once again for most metrics, suggesting that domain adaptation improves generalization for unseen centers.

\subsection{Varying the consistency loss}
We executed multiple runs of the Mean Teacher algorithm by varying the consistency loss to determine which one works best. We focused just on losses that do not contain additional hyperparameters. The Tversky Loss~\cite{salehi2017tversky}, for instance, is quite similar to the Dice loss but with two additional hyperparameters ($\alpha$ and $\beta$). 

Our choices of losses were thus limited to cross-entropy, mean squared error (MSE), and Dice, as previously described in Section~\ref{sec:method}. We believe, however, that a thorough analysis of distinct loss functions is of great importance for domain adaptation and should be explored in future work.

\subsection{Behavior of Dice loss and thresholding}
A well-known fact regarding the Dice loss is that it usually produces predictions concentrated around the upper and lower bounds of the probability distribution, with very low entropy. As in~\cite{Perone2018}, we used a high threshold value ($0.99$) for the Dice predictions to produce a balanced model. We have found, however, that the domain adaptation method also regularizes the network predictions, shifting the Dice probability distribution outside of the probability bounds. For that reason, we have decreased the Dice prediction threshold to 0.9 (instead of 0.99), which produced a more balanced model in terms of precision and recall.

\subsection{Training stability}
For unsupervised domain adaptation, it is important to have a stable training procedure. Since, in the most difficult scenarios, there are no annotations for validating the adaptation, an unstable training may produce sub-optimal adaptation results.

\begin{table*}[!htpb]
    \centering
    \caption{Results on evaluating on center 3. The training set includes centers 1 and 2 simultaneously, with unsupervised adaptation for center 3. Values within parentheses represent the best validation results for each metric. The remaining values represent the final result after 350 epochs.}
    \label{tab:unstable}
    \begin{tabular}{ccccccccc}
        \toprule
        \textbf{Loss} & \textbf{Weight} & \textbf{Dice} & \textbf{mIoU} & \textbf{Recall} & \textbf{Precision} & \textbf{Specificity} & \textbf{Hausdorff} \\
        \midrule
        \multirow{4}{*}{CE} & 5 & 0.00 (85.50) & 0.00 (74.91) & 0.00 (95.01) & 0.00 (98.90) & 100.0 (100.00) & 0.00 (0.00) \\
        & 10 & 0.00 (80.73) & 0.00 (69.54) & 0.00 (83.21) & 0.00 (98.78) & 100.0 (100.00) & 0.00 (0.00) \\
        & 15 & 6.43 (37.03) & 4.89 (26.06) & 5.38 (77.05) & 17.34 (65.85) & 100.0 (100.00) & 0.28 (0.00) \\
        & 20 & 2.30 (67.61) & 1.86 (52.55) & 2.09 (65.00) & 7.94 (96.57) & 100.0 (100.00) & 0.12 (0.03) \\ 
        
        \midrule
        
        \multirow{4}{*}{Dice} & 5 & 76.76 (80.74) & 62.76 (68.16) & 97.88 (99.66) & 63.72 (72.50) & 99.71 (99.81) & 2.36 (2.16) \\
        & 10 & 4.77 (10.55) & 2.45 (5.64) & 96.25 (99.99) & 2.45 (5.85) & 79.59 (99.75) & 8.80 (2.57) \\
        & 15 & 2.30 (7.74) & 1.16 (4.12) & 99.95 (100.00) & 1.16 (4.62) & 55.07 (99.80) & 11.75 (2.50) \\
        & 20 & 1.79 (4.43) & 0.90 (2.27) & 99.99 (100.00) & 0.90 (2.30) & 42.02 (99.84) & 12.68 (2.43) \\ 
        
        \midrule
        
        \multirow{4}{*}{MSE} & 5 & 83.7 (83.88) & 72.2 (72.46) & 91.24 (98.19) & 78.1 (78.57) & 99.87 (99.93) & 2.1 (2.00)\\
        & 10 & 84.38 (84.38) & 73.19 (73.19) & 90.15 (99.07) & 80.12 (80.12) & 99.88 (99.94) & 2.05 (1.89)\\
        & 15 & 84.59 (84.59) & 73.49 (73.50) & 89.19 (98.52) & 81.28 (81.28) & 99.89 (99.89) & 2.03 (2.03)\\
        & 20 & 84.5 (84.50) & 73.36 (73.37) & 90.36 (94.63) & 80.16 (80.16) & 99.88 (99.98) & 2.05 (1.46)\\

        \bottomrule
    
    \end{tabular}
\end{table*}

To evaluate the training stability, we tried distinct consistency weights for each possible consistency loss and we evaluated the difference between the best values that were found and the final results after 350 epochs. Table~\ref{tab:unstable} summarizes results of this analysis.

\begin{figure}[!htpb]
	\centering
	\includegraphics[width=1.0\linewidth]{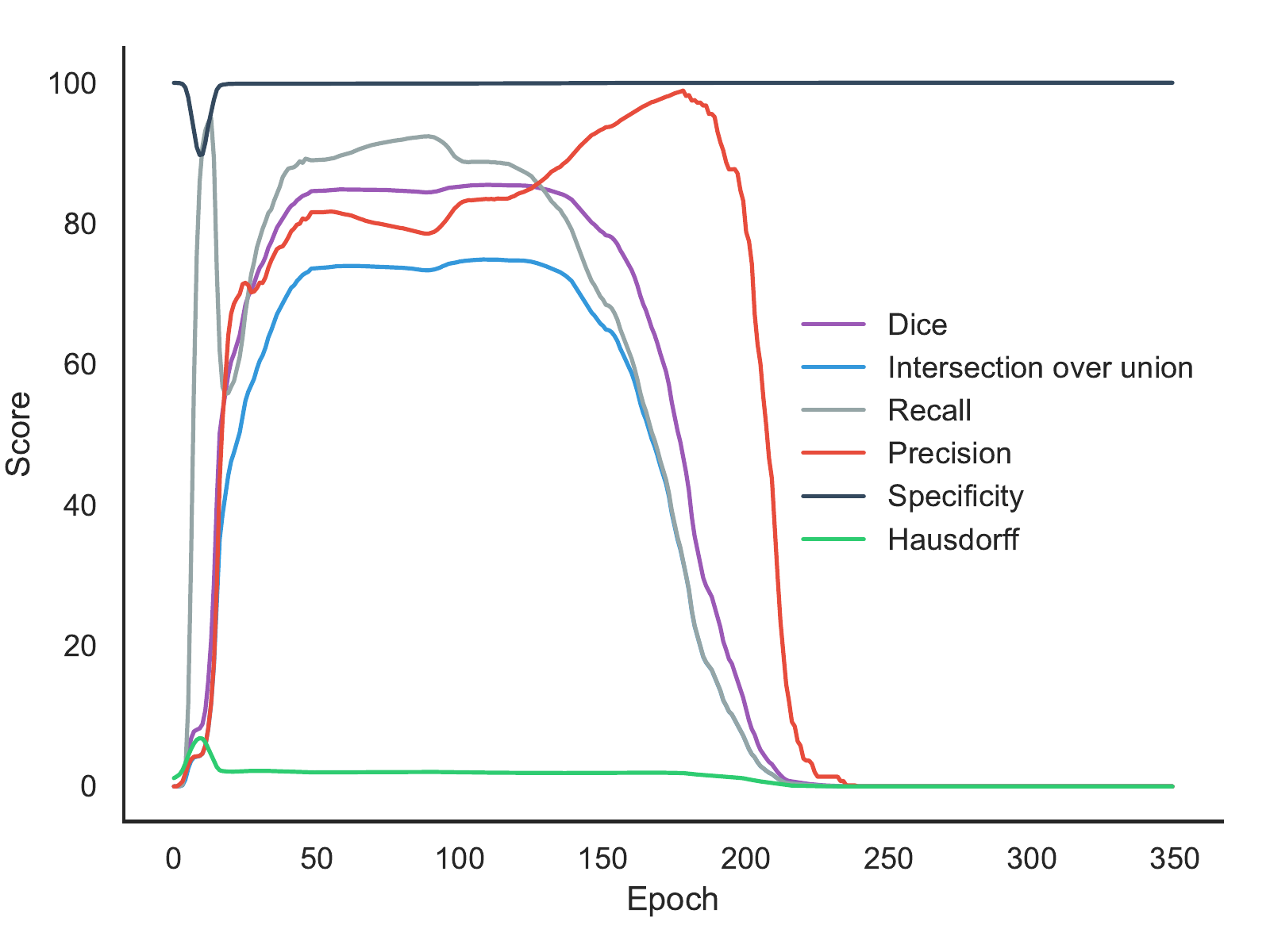}
	\caption{\label{fig:unstable}
        Per-epoch validation results for the teacher model at center 3 with cross-entropy as the consistency loss. Training was conducted in both centers 1 and 2 simultaneously, and adapted to center 3 with consistency weight $\gamma = 5$. Best viewed in color.
	}
\end{figure}

We can observe that cross-entropy consistently fails, even with different weights, potentially due to the class imbalance of this particular task. Though it also achieves high dice values in its best scenario during training. Thus cross-entropy becomes a possible alternative to MSE when a few annotated images are available for validation in the target domain. Figure~\ref{fig:unstable} shows how the training diverges for cross-entropy after several iterations.

We can observe that both Dice and cross entropy have trouble stabilizing the training after achieving high results. However, MSE tends to be more invariant to consistency weight, thus being a robust approach when no annotated data is available at the target center. As in \cite{french2017self}, we also tried confidence thresholding, although we did not observe improvements.

\section{Ablation studies}
\label{sec:ablation-experiments}
This section describes the ablation analyses, the purpose of which was to better understand the behavior of different components in the domain adaptation scenario.

\subsection{Exponential moving average (EMA)}
The improvement seen in Table~\ref{tab:results} could also be explained by introducing the exponential moving average (EMA) during the training procedure, since it averages and smoothes the SGD trajectories.

To demonstrate that the improvement is specific to using unlabeled data and does not only come from the exponential average component, we performed an ablation experiment that leaves the EMA active but sets the consistency weight to zero. This experiment allowed us to evaluate the impact of the exponential average in the absence of the unlabeled data used to enforce consistency.

We reproduced the same experimental setup from the Table~\ref{tab:results} but with the consistency weight set to zero. Results are presented in Table~\ref{tab:polyak} and show that the EMA model (teacher) presents no gains over the non-averaged model (the supervised baseline). This could arguably be due to a poorly chosen $\alpha$. However, note that Mean Teacher, which heavily relies on the EMA model, was nevertheless able to outperform a purely-supervised method by a great margin as seen in Table~\ref{tab:results}.

\begin{table*}[!htpb]
    \centering
    \caption{Results of the ablation experiment where the baseline model was trained and compared
against its exponential moving average (EMA) model without using Mean Teacher training scheme with unlabeled data. All experiments were trained in both center 1
and 2 simultaneously. Center 3 is the validation set and Center 4 is the test set.}
    \label{tab:polyak}
    \begin{tabular}{cccccccc}
        \toprule
        \textbf{Evaluation} & \textbf{Version} & \textbf{Dice} & \textbf{mIoU} & \textbf{Recall} & \textbf{Precision} & \textbf{Specificity} & \textbf{Hausdorff} \\ \midrule
        

        \multirow{2}{*}{Center 3} & Baseline & 83.06&	71.36&	90.98&	77.24&	99.86	&2.13 \\
        & EMA & 83.09&	71.40&	90.97&	77.30&	99.86&	2.13 \\
        \midrule

        \multirow{2}{*}{Center 4} & Baseline & 69.41&	53.90&	97.20&	54.98&	99.70	&2.48 \\
        & EMA & 69.50&	54.00&	97.19&	55.09&	99.71	&2.48 \\
    \bottomrule
    \end{tabular}
\end{table*}

\begin{figure*}[!htpb]
	\centering
    \begin{subfigure}{.5\textwidth}
    	\captionsetup{margin=.5cm}
    	\includegraphics[width=\textwidth]{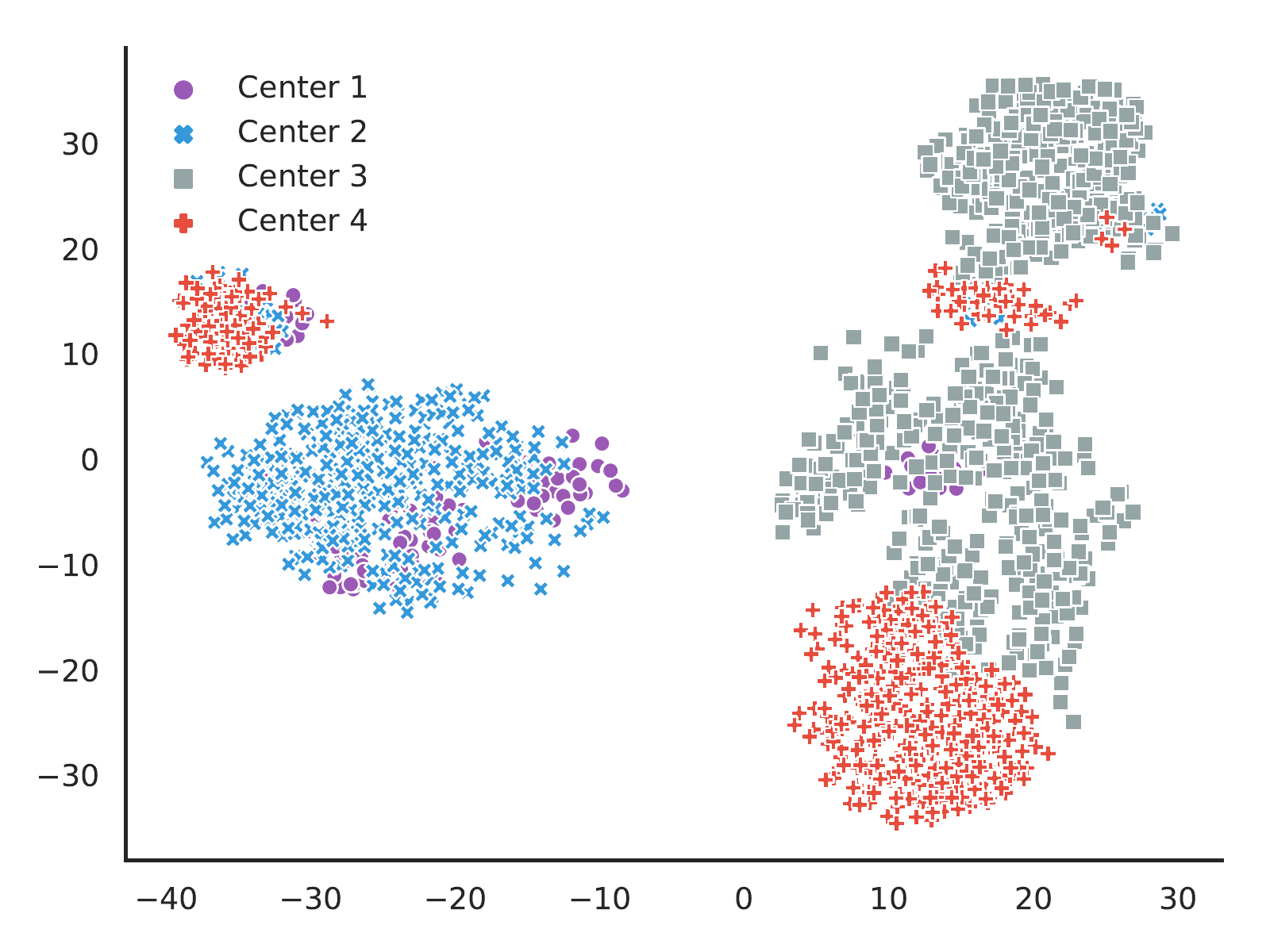}
    	\caption{\label{fig:tsne_supervised}
    	    A visualization of the t-SNE 2D non-linear embedding projection for the supervised learning scenario.
    	    The colors represent data from different centers.
    	}
    \end{subfigure}%
    \begin{subfigure}{.5\textwidth}
    	\captionsetup{margin=.5cm}
    	\includegraphics[width=\textwidth]{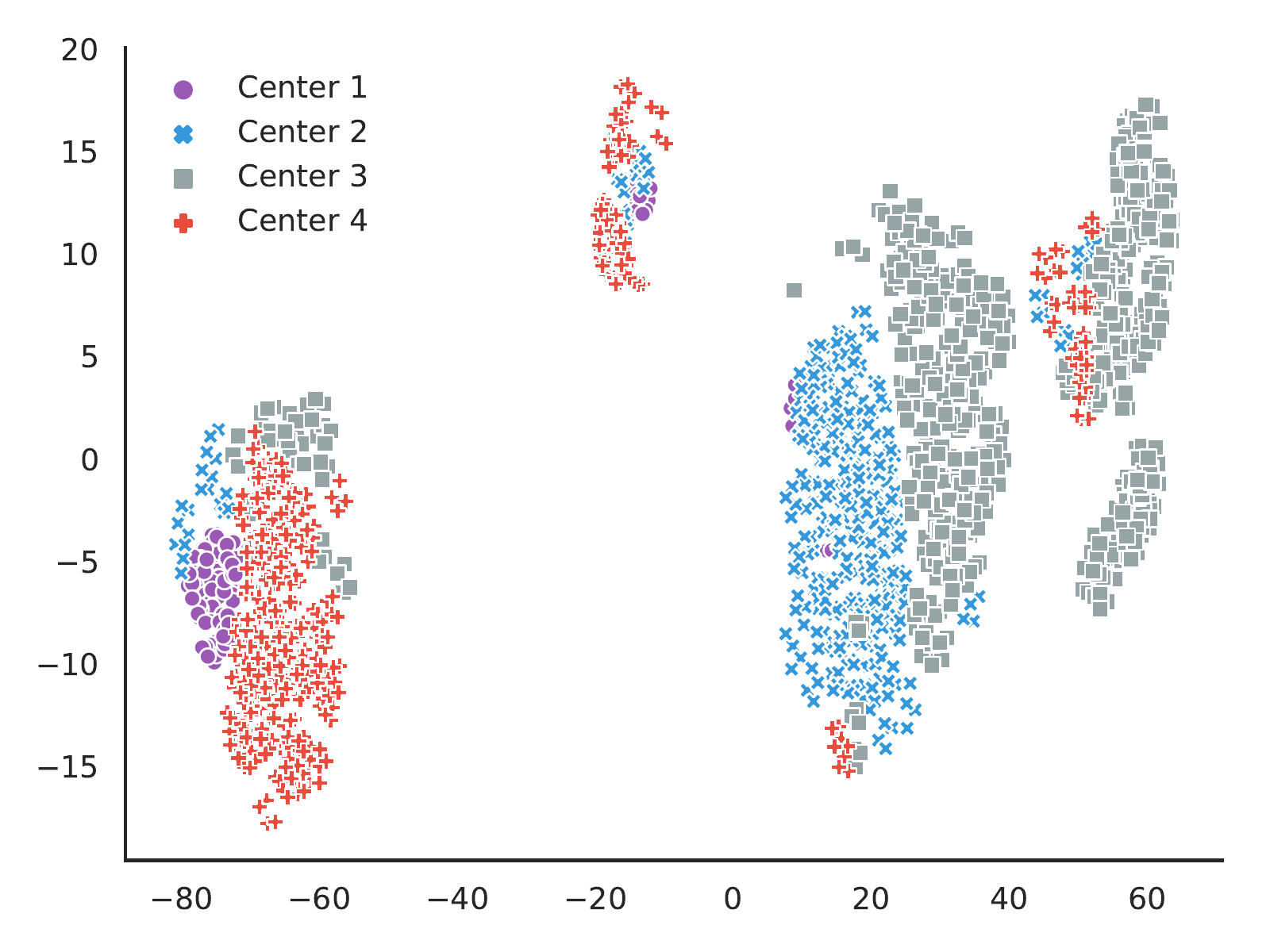}
    	\caption{\label{fig:tsne_adapt}
    	    A visualization of the t-SNE 2D non-linear embedding projection for the domain adaptation scenario.
    	    The colors represent data from different centers.
    	}
    \end{subfigure}
    \caption{Execution of t-SNE algorithm for two different scenarios. Best viewed in color.}
\end{figure*}

\section{Domain shift visualization}
\label{sec:visualization}

Next, we investigated how domain adaptation affects the prediction space of segmentation at distinct centers. By using $t$-SNE~\cite{maaten2008visualizing}, a non-linear dimensionality reduction technique, we were able to assess changes on the predictive perception of the network regarding unsupervised data. All data presented in the following figures were not used for training.

We created two baselines for this experiment. The first model was trained in a supervised fashion following the same hyperparameters presented in Section~\ref{sec:baseline}. The second was an adaptation scenario where both centers 1 and 2 were used as supervised centers and 3 as adaptation target. The vectors projected with $t$-SNE represents the features from the network prior to the final sigmoid activation.

Both $t$-SNE executions had a learning rate set to 10, perplexity to 30, and were executed for about 1,000 iterations\footnote{We used the TensorBoard embedding projector, available at \url{https://github.com/tensorflow/tensorboard}}. We notice that more iterations than 1,000 preserved the groups' structure but further compressed them. This made visualizing the centers harder, so 1,000 was a good trade-off between identifying emerging groups and interpretability. 

Results from the supervised experiment are shown in Figure~\ref{fig:tsne_supervised}. Note that there is a clear separation between data from centers used during training (1 and 2) and unseen centers (3 and 4). This shows that the network predictions greatly differ according to the center to which the sample belongs to.

When adapting the network with unlabelled samples from a different domain, predictions become more diffuse, at least for centers presented during training. Results from the unsupervised adaptation experiment are shown in Figure~\ref{fig:tsne_adapt}. In that scenario, centers with labeled data (centers 1 and 2) form clusters with domains seen only in an unsupervised manner (3) or not presented to the network at all (4). A possible explanation for the close proximity of clusters (1, 4) and of clusters (2, 3) is the similarity of intensity distribution within each pair of clusters, as highlighted in Figure~\ref{fig:distribution}. See appendix for more details regarding the relationship between the data distribution and the t-SNE clusters.

\section{Conclusion and limitations}
\label{sec:conclusion-limitation}
Variability and scarcity of annotations in the medical imaging context is still challenging for machine learning. The large set of parameters that can be used to acquire image modalities and the lack of standardized protocols or industry standards are pervasive across the entire field.

In this work, we have shown that unsupervised domain adaptation, without depending on annotations, is an effective way to increase the performance of machine learning models for medical imaging across multiple centers.

Through the evaluation of multiple metrics in a large set of experiments, we have shown how self-ensembling methods can improve generalization on unseen domains through the leverage of unlabeled data from multiple domains. We also performed an ablation study that demonstrated strong evidence that the improvements come by the introduction of the unlabeled data and not only due to the exponential moving average.

We assessed how cross-entropy (when used as a consistency loss function) fails at maintaining training stability when the number of epochs progresses. We have discussed how this can lead to potential problems in more challenging scenarios for multiple centers. We also discussed issues related to the Dice loss when used as consistency loss.

We acknowledged the following limitations in our study. Firstly, we did not evaluate adversarial training methods for domain adaptation. Even considering the Mean Teacher as the current state-of-the-art method on many datasets, we believe that further analyses on the same realistic small data regime could significantly increase the importance of our contributions, and thus we leave that aspect for future work. 

Secondly, the single-task evaluation of the gray matter segmentation could be extended to other tasks in other domains. Increasing the number of centers alongside the number of tasks would be relevant for confirming results obtained in the present study. 

Further work on the field could lead to methods capable of measuring the risk of adaptation to particular centers or domains. This would be an important step towards understanding the limitations of the domain adaptation methods. 

We believe that the problems that arise from the variability of medical imaging modalities require rethinking some of the strong assumptions made in machine learning models and training procedures. An important step in that direction is to reassess the importance of proper multi-domain evaluation in studies and medical imaging challenges, which rarely provide a test set from different domains (such as different centers, machines, etc) that contain the variability found in real-world scenarios.

\section{Source-code and dataset availability}
\label{sec:source-code}
In the spirit of Open Science and reproducibility, the source-code used to perform the experiments presented in this study is publicly available~\footnote{https://github.com/neuropoly/domainadaptation}.

The dataset used for this work is also available on the Spinal Cord Gray Matter Segmentation Challenge website\footnote{http://cmictig.cs.ucl.ac.uk/niftyweb/program.php?p=CHALLENGE}.

\section{Acknowledgments}
\label{sec:acknowledgments}
We are very thankful to Ryan Topfer for the sensible review and time dedicated to improve this article. Funded by the Canada Research Chair in Quantitative Magnetic Resonance Imaging [950-230815], the Canadian Institute of Health Research [CIHR FDN-143263], the Canada Foundation for Innovation [32454, 34824], the Fonds de Recherche du Québec - Santé [28826], the Fonds de Recherche du Québec - Nature et Technologies [2015-PR-182754], the Natural Sciences and Engineering Research Council of Canada [435897-2013], the Canada First Research Excellence Fund (IVADO and TransMedTech) and the Quebec BioImaging Network [5886]. This study was financed in part by the Coordenação de Aperfeiçoamento de Pessoal de Nivel Superior – Brasil (CAPES) – Finance Code 001.

\bibliography{references}

\appendix
\section{Extended visualizations}
\begin{figure*}[!htpb]
	\centering
	\includegraphics[width=\linewidth]{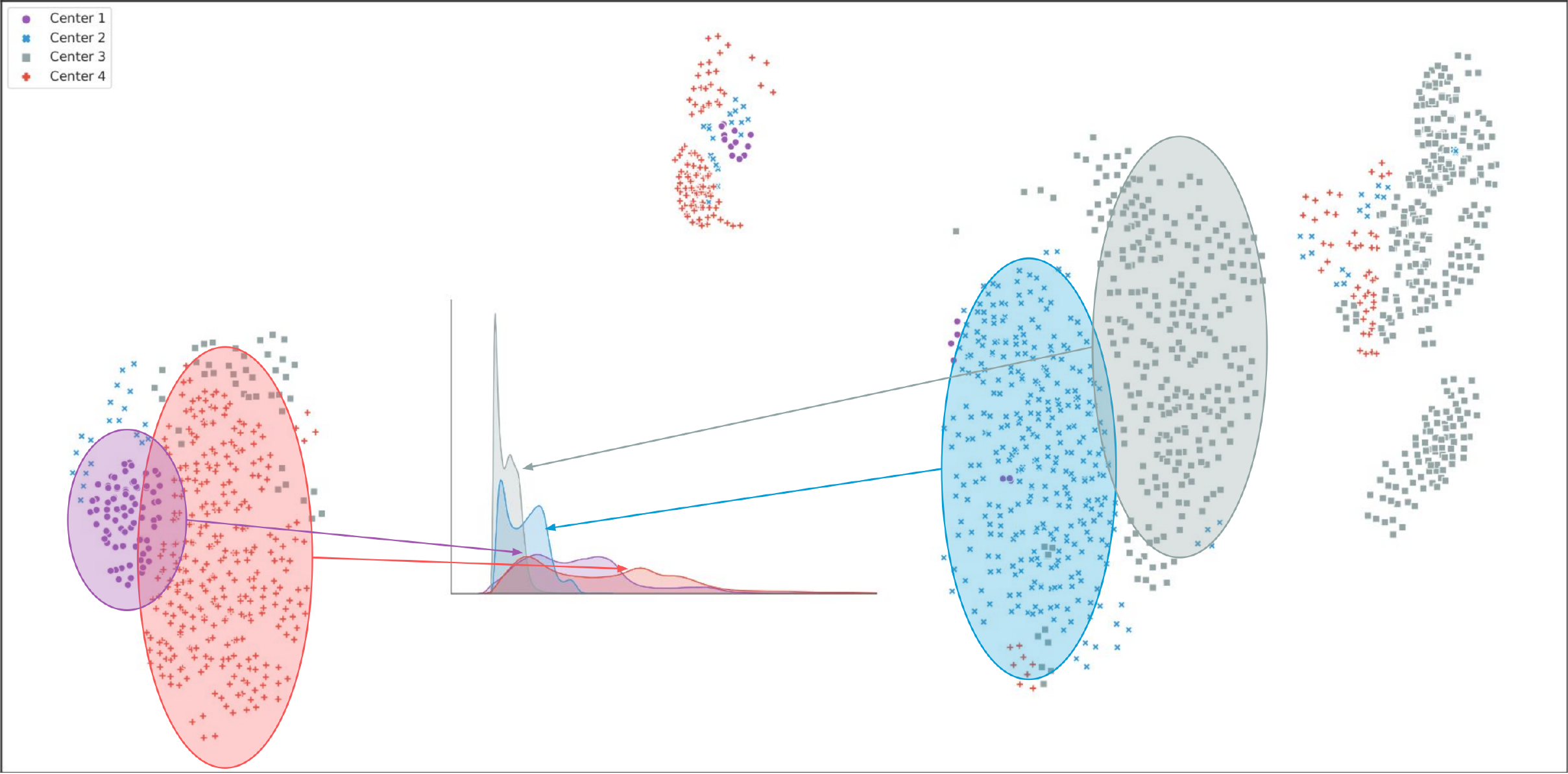}
	\caption{\label{fig:clusters_tsne}
        Extended visualization based on $t$-SNE embedding from the domain adaptation scenario in Figure~\ref{fig:tsne_adapt}. The chart in the middle represents the pixel distribution from each center. Note how similar distributions tend to form clusters on the prediction space. Best viewed in color.
	}
\end{figure*}

In Figure~\ref{fig:clusters_tsne} we show an extended visualization of the t-SNE embeddings from the domain adaptation scenario where the underlying raw intensity distribution is described together with their respective clusters.

\end{document}